\documentclass[runningheads]{llncs}

\usepackage[T1]{fontenc}
\usepackage{graphicx}
\usepackage{float}
\usepackage{multirow}
\usepackage[table,xcdraw]{xcolor}
\usepackage[hidelinks]{hyperref}
\usepackage{subcaption}
\usepackage{color}
\usepackage{array}
\usepackage{booktabs}

\usepackage{caption}
\begin{document}

\title{On Biases in a UK Biobank-based Retinal Image Classification Model}

\author{Anissa Alloula\inst{1}\orcidID{0000-0003-1525-3994}
% index{Last Name, First Name}  
\and Rima Mustafa\inst{1}\orcidID{0000-0003-2623-5337}
% index{Last Name, First Name} 
\and
Daniel R McGowan\inst{2,3}\orcidID{0000-0002-6880-5687} 
% index{Last Name, First Name} 
\and Bart\l omiej W. Papie\.z\inst{1}\orcidID{0000-0002-8432-2511}
% index{Last Name, First Name} 
}
\authorrunning{A. Alloula et al.}
\institute{Big Data Institute, University of Oxford, United Kingdom \\
\email{anissa.alloula@dtc.ox.ac.uk}
\and
Department of Oncology, University of Oxford, United Kingdom \and Department of Medical Physics and Clinical Engineering, Oxford University Hospitals NHS FT, Oxford, United Kingdom.}
\maketitle
\begin{abstract}
Recent work has uncovered alarming disparities in the performance of machine learning models in healthcare. In this study, we explore whether such disparities are present in the UK Biobank fundus retinal images by training and evaluating a disease classification model on these images. We assess possible disparities across various population groups and find substantial differences despite strong overall performance of the model. In particular, we discover unfair performance for certain assessment centres, which is surprising given the rigorous data standardisation protocol. We compare how these differences emerge and apply a range of existing bias mitigation methods to each one. A key insight is that each disparity has unique properties and responds differently to the mitigation methods. We also find that these methods are largely unable to enhance fairness, highlighting the need for better bias mitigation methods tailored to the specific type of bias.

\end{abstract}

\section{Introduction and Related Work}

\subsubsection{Biases and Disparities in Machine Learning.}
An emerging concern in machine learning (ML) research is that strong overall performance may obscure critical disparities, leading to substantially inferior outcomes for certain subgroups. Examples of this unequal performance have been identified in clinical ML models, across a range of tasks and modalities such as skin lesion classification \cite{bevanSkinDeepUnlearning2022}, brain Magnetic Resonance Imaging (MRI) reconstruction \cite{duUnveilingFairnessBiases2023}, cardiac MRI segmentation \cite{puyol-antonFairnessCardiacMR2021}, and affecting various subgroups, from certain ethnic groups \cite{puyol-antonFairnessCardiacMR2021,kumarDistributionallyRobustOptimization2023}, to disadvantaged socioeconomic groups \cite{seyyed-kalantariUnderdiagnosisBiasArtificial2021}. Not only do these biases harm the minority groups who are subject to them, but they also hinder the generalisability of the models to unseen population samples \cite{sanchezCausalMachineLearning2022}, constituting a major barrier to the implementation of ML models in clinical settings. 

\subsubsection{Existing Approaches to Address such Biases}
A line of research focused on preventing such disparities has consequently emerged. Bias mitigation can be conducted at various stages in the ML pipeline: during data collection, in the pre-processing stage, while the model is training, and/or in post-processing. Objectives vary between methods and can include boosting minimum performance \cite{dianaMinimaxGroupFairness2021}, reducing gaps in performance (equalised odds \cite{duUnveilingFairnessBiases2023}), or equalising the number of positive predictions across groups (demographic parity \cite{castelnovoClarificationNuancesFairness2022}). However, recent work has highlighted that despite the multitude of existing methods, the problem is far from solved. A benchmark from 2023, MEDFAIR, showed that across a range of medical tasks, no method consistently and significantly outperformed empirical risk minimisation (where there is no fairness objective) \cite{zongMEDFAIRBENCHMARKINGFAIRNESS2023}. 

\subsubsection{Problem Setting}
In this study, we focus on the appearance of biases and their mitigation in retinal imaging-based models. Bias mitigation research has been lacking in this field, with, to the best of our knowledge, only two examples: work by Burlina et al. \cite{burlinaAddressingArtificialIntelligence2021} and work by Coyner et al. \cite{coynerAssociationBiomarkerBasedArtificial2023}, who tried to mitigate race-related disparities with synthetic data and data pre-processing, respectively. We build on this work by conducting the largest and most comprehensive exploration of disparities and mitigation methods in retinal imaging to date. We use retinal images from the UK Biobank (UKBB), an unparalleled medical database of over half a million UK adults \cite{sudlowUKBiobankOpen2015}. We complement recent work which has identified selection bias in the UKBB  \cite{lyallQuantifyingBiasPsychological2022,swansonUKBiobankSelection2012,bradleyAddressingSelectionBias2022,schoelerParticipationBiasUK2023} by considering other possible bias types and how they manifest in ML models. In addition to providing insights on understudied possible biases in retinal imaging, the use of this database allows us to consider what disparities remain when standardisation has been conducted, as the UKBB has undergone rigorous data acquisition and quality control protocols \cite{allenProspectiveStudyDesign2024}, such that all images were taken with the same type of OCT scanner \cite{Resource100237}. Also, the breadth of data available in the UKBB allows us to specifically characterise different biases (including some which are rarely investigated). 

\subsubsection{Contributions}

We train a retinal image hypertension classification model on images from more than 75,000 individuals and use this as a proxy task to understand possible biases. We find that our model has uneven performance across subgroups, including between images from different assessment centres. We explore possible reasons for these disparities among common factors such as data imbalance, image quality, unequal generalisation, and separations in the model's representations of different subgroups, and find that these do not necessarily hold true depending on the disparity. Finally, we find that no bias mitigation method manages to consistently improve the fairness of our model. This highlights the non-universality of existing bias mitigation methods and underscores the need for a framework to specifically characterise disparities and their causes, as well as to determine if and how to best minimise them.  

\section{Methods and Experimental Setup}

\subsubsection{Dataset and Pre-Processing}

We use 80,966 fundus retinal images from the right eye of 78,346 individuals in the UKBB. We exclude 1,874 images corresponding to participants who had subsequently withdrawn, who had ``other'', ``preferred not to say'', or ``unknown'' ethnicity, and those from one assessment centre which had fewer than 0.2\% of images. 
The UKBB is particularly rich in available metadata, including age, body mass index (BMI), self-reported alcohol consumption, self-reported ethnicity, genetic ethnicity (gen\_ethnicity), genetic sex, deprivation, medication, etc. We create categorical groupings for age (40-50, 50-60, 60-70, 70+), BMI (0-3 based on quartile), deprivation index (0-3 based on quartile), and self-reported ethnicity (White, mixed background, Asian background, or Black African background) to facilitate downstream analyses. We anonymise the names of the centres.

We also adjust diastolic and systolic blood pressure (BP) by +10 and +15~mm~Hg, respectively, if individuals are taking hypertensive medication \cite{tobinAdjustingTreatmentEffects2005}. We classify individuals as having high blood pressure (hypertension) if: diastolic BP $\geq$ 80 or systolic BP $\geq$ 130 or if they are taking anti-hypertensive medication (according to the current guidelines \cite{whelton2017ACCAHA2018}). This is the binary target variable our model aims to predict. Figure A1 shows some of the dataset characteristics.

\subsubsection{Model Architecture and Training}

We split data into train, validation, and test sets (0.8, 0.1, 0.1) stratifying by individuals. As in \cite{poplinPredictionCardiovascularRisk2018}, we train an InceptionV3 Network (\cite{szegedy2015rethinking}) to classify a retinal image as belonging to a hypertensive or non-hypertensive individual. Table \ref{Implementation table} shows specific implementation details.

\begin{table}[htbp]
\centering
\caption{Implementation details for UKBB Implementation.}\label{Implementation table}
\resizebox{10cm}{!}{%
\begin{tabular}{ll}
\toprule
\textbf{Training strategy}    & \textbf{UKBB Implementation}         \\ \midrule
Network backbone              & InceptionV3                                  \\ \hline
Pre-training          & ImageNet                                   \\ \hline
Batch size                  & 512                                     \\ \hline
Image size                   & 3x299x299                           \\ \hline
Augmentation                  & Random flip, rotation, crop, color jitter, Gaussian blur                                     \\ \hline
Optimiser                    & Adam                                     \\ \hline
Loss & Binary cross-entropy                                  \\ \hline
Learning rate                & 0.0005             \\ \hline
Learning scheduler               & StepLR (gamma = 0.1, step = 10)                                         \\ \hline
Weight decay                 & 0.0001             \\ \hline
Max epochs                 & 100 (with early stopping after 10) \\ \bottomrule
\end{tabular}%
}
\end{table}

\subsubsection{Bias Mitigation Models}

We adapt implementations of existing bias mitigation methods from the github repository \href{https://github.com/ys-zong/MEDFAIR}{MEDFAIR}, using the same backbone and core parameters as in Table \ref{Implementation table}. We select methods which encompass different types of bias mitigation approaches and which had good results in the MEDFAIR benchmark \cite{zongMEDFAIRBENCHMARKINGFAIRNESS2023} and try to mitigate age-, assessment-centre-, and sex-related disparities.

We test \textbf{Resampling} of minority subgroups as a pre-processing method \cite{idrissiSimpleDataBalancing2022}. In addition, we explore a range of in-processing methods including \textbf{Group Distributionnally Robust Optimisation (GroupDRO)} which minimises worst-group loss \cite{sagawaDistributionallyRobustNeural2020,kumarDistributionallyRobustOptimization2023}, \textbf{Orthogonally Disentangled Representations (ODR)}, which disentangles the representations of subgroup-related features and task-relevant features \cite{sarhanFairnessLearningOrthogonal2020}, \textbf{Domain-Independent learning (DomainInd)} where each subgroup has its own final classification layer \cite{wangFairnessVisualRecognition2020}, and \textbf{Learning-Not-to-Learn (LNL)}, an adversarial learning method \cite{kimLearningNotLearn2019}. We also implement \textbf{Stochastic Weight Averaging Densely (SWAD)} \cite{chaSWADDomainGeneralization2021} which is a general robustness method (and therefore does not require subgroup information) and pair it with resampling (\textbf{ReSWAD}). Finally, we implement a post-processing method (not in MEDFAIR), \textbf{Recalibration}, where a different decision threshold is calculated for each subgroup. We train all models three times with different random seeds on NVIDIA A100 GPU's.

\subsubsection{Model Evaluation}

Model evaluation is based upon the mean Receiver Operating Characteristic Area Under the Curve (AUC), accuracy, precision, and recall scores for the three runs of each model. We consider overall performance and performance across different subgroups (both minimum performance and best- and worst- performance gap). All code is available at \url{https://github.com/anissa218/MEDFAIR_UKBB}.

\section{Results and Discussion}

\subsubsection{Performance and Disparities of the Baseline Model}
\begin{figure}[htbp]
    \centering
    \resizebox{10cm}{!}%{\includegraphics{baseline_model_figs/baseline_disparities_plot_adjusted_labels.png}}
    {\includegraphics{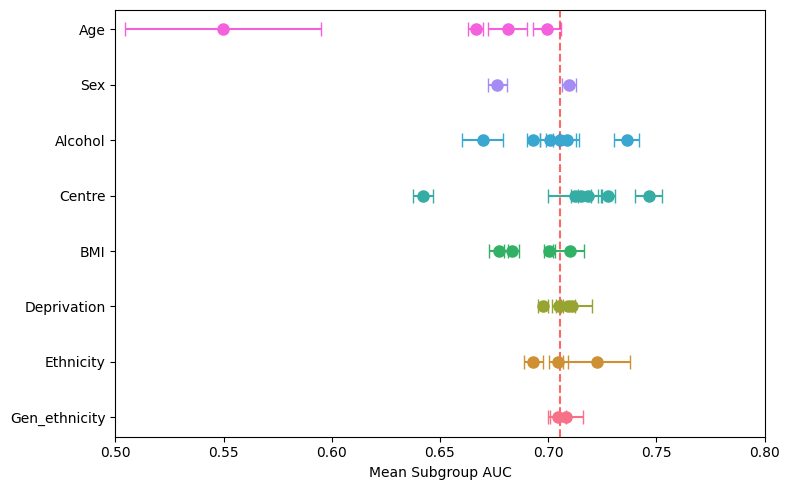}}
    \caption{For some subgroupings, the baseline model shows large disparities in test set AUC between worst- and best- performing subgroups, far below and above the average AUC of 0.71. Error bars represent standard deviation across the three random seeds.}
    \label{Plot overall gaps and min performance}
\end{figure}

The baseline InceptionV3 model achieves 73$\pm$0.01\% accuracy and 71$\pm$0.00\% AUC in hypertension classification, with precision and recall values of 81$\pm$0.04\% and 83$\pm$0.01\%, respectively. 
However, a more granular assessment reveals significant disparities across certain subgroups (Table A1). For instance, as shown in Figure \ref{Plot overall gaps and min performance}, the model's AUC varies by over 15\% between different age groups and 10\% between centres, with the worst-group AUC being substantially lower than the average AUC of 0.71. Some subgroups also exhibit substantial differences in recall (which would translate to underdiagnosis) of 10 to 32\%, including different age groups, assessment centres, alcohol consumers, and ethnic groups. 

\subsubsection{Origins of Performance Disparities}

Next, we aim to understand why these disparities appear. We investigate whether they can be attributed to varying underlying characteristics across subgroups, such as differences in age or sex distribution. However, regardless of the attribute we condition on, the worst-performing assessment centre, centre f, shows much lower AUC (results on age conditioning are shown in Table A2). Such trends are also preserved for sex- and age-related disparities. Additionally, we use the Automorph pipeline \cite{automorph} to assess the quality of all images and use this as a conditioning variable. We find that image quality does not explain these disparities either. We also consider shifts in prevalence, as correlation between an attribute and the target label can cause bias \cite{jonesCausalPerspectiveDataset2024}. This is evident in age- and sex-related disparities, where hypertension shows a strong positive correlation with age (Figure A1), and men have a higher prevalence of hypertension. However, this does not explain centre disparities, as the worst-performing subgroup has approximately 76\% images with hypertension, which falls within the range of other centres (69\%-80\%).

Further, these differences cannot simply be attributed to data imbalance. For centre and sex-related disparities, all groups are evenly represented. However, for age-related disparities, data imbalance may play a role. The oldest age group, which has the lowest AUC, is also underrepresented, comprising only 2.5\% of the images.

Another emerging hypothesis in fair ML research is that disparities arise due to unequal model generalisation across subgroups. Despite uniform and strong performance on training data, generalisation differences on unseen data can emerge \cite{duttFairTuneOptimizingParameter2024,sagawaDistributionallyRobustNeural2020}. As shown in Table \ref{Table trainvaltest}, there is a noticeable decrease in worst-group AUC relative to the decrease in overall AUC between training data and test data for different centres. Similarly, the gap between centres increases on unseen data, suggesting that the model's generalisation varies across these centres. The difference is not as striking for age and sex subgroups, and most likely simply linked to overall performance decrease on unseen data. We further investigate whether there is a shift in generalisation during training; a point where the model starts overfitting to certain subgroups but not others (and thus increasing the gap between subgroups) as identified in \cite{duttFairTuneOptimizingParameter2024}. However our analyses do not reveal any evidence of a specific point where this could occur (Figures A2).

\begin{table}[htbp]
\centering
\caption{Age, centre, and sex disparities across seen and unseen data (Test AUC - Train AUC). While disparities increase in unseen test data for all groups, the increase is strongest for assessment centres, suggesting unequal generalisation. Standard deviation of the three random seeds shown in parentheses.}
\resizebox{10cm}{!}{%
\begin{tabular}{>{\centering\arraybackslash}p{2.6cm}>{\centering\arraybackslash}p{2.6cm}>{\centering\arraybackslash}p{2.6cm}>{\centering\arraybackslash}p{2.6cm}}
\toprule
\textbf{Subgroup} & \textbf{$\Delta$ Overall AUC} & \textbf{$\Delta$ Min AUC} & \textbf{$\Delta$ AUC Gap} \\ \midrule
\textbf{Age}      & -0.031 (0.011)             & -0.037 (0.055)                 & 0.004 (0.044)                \\ \hline
\textbf{Centre}   & -0.031 (0.011)             & -0.045 (0.014)                 & 0.032 (0.006)                \\ \hline
\textbf{Sex}      & -0.031 (0.011)             & -0.034 (0.011)                 & 0.009 (0.002)                \\ \bottomrule
\end{tabular}%
}
\label{Table trainvaltest}
\end{table}

\begin{figure}[htbp]
    
    \resizebox{12cm}{!}
    {\includegraphics{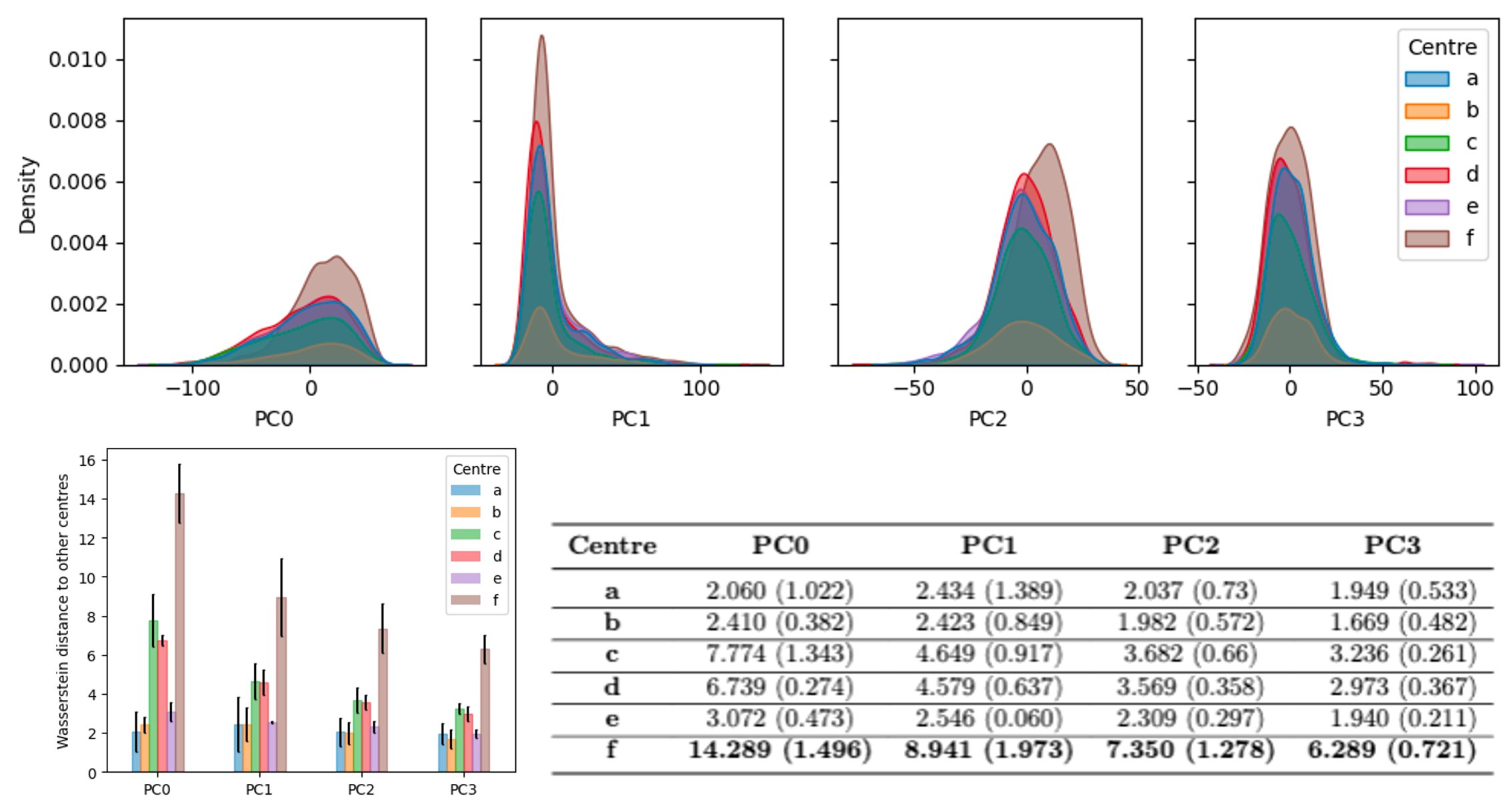}}
    %{\includegraphics{baseline_model_figs/baseline44_kde_centre.png}}
    \caption{Kernel density estimation of the first 4 principal components (PC) of the features extracted from the baseline model's penultimate layer grouped by centre. Table of mean Wasserstein distance of features between one centre and the other 5 for the 3 random seeds. f's feature distribution is clearly an outlier across some PCs.}
    \label{centre features kde}
\end{figure}

Finally, we investigate whether the model’s learnt representations can provide insight on subgroup disparities. We analyse each image in the model’s penultimate layer feature space through a 4-component principal component analysis (which explains over 85\% of the variance). As expected, we find strong separation between the projected features of images with and without hypertension, and consequently between images of different age groups due to their strong correlation. However, we also observe an unexpected outlier from the distribution of images from the worst-performing centre (f). There is a clear difference in the kernel density estimates of some principal components from this centre and a consistently increased Wasserstein distance separating the distribution of features from centre f to the other centres (Figure \ref{centre features kde}). Although this does not prove this information is being used for predictions, it is noteworthy that such a shift exists, one that cannot be explained by any of the other available variables. 

\subsubsection{Overall Performance of Mitigation Models}

\begin{figure}[htbp]
    \centering
    \resizebox{12cm}{!}{\includegraphics{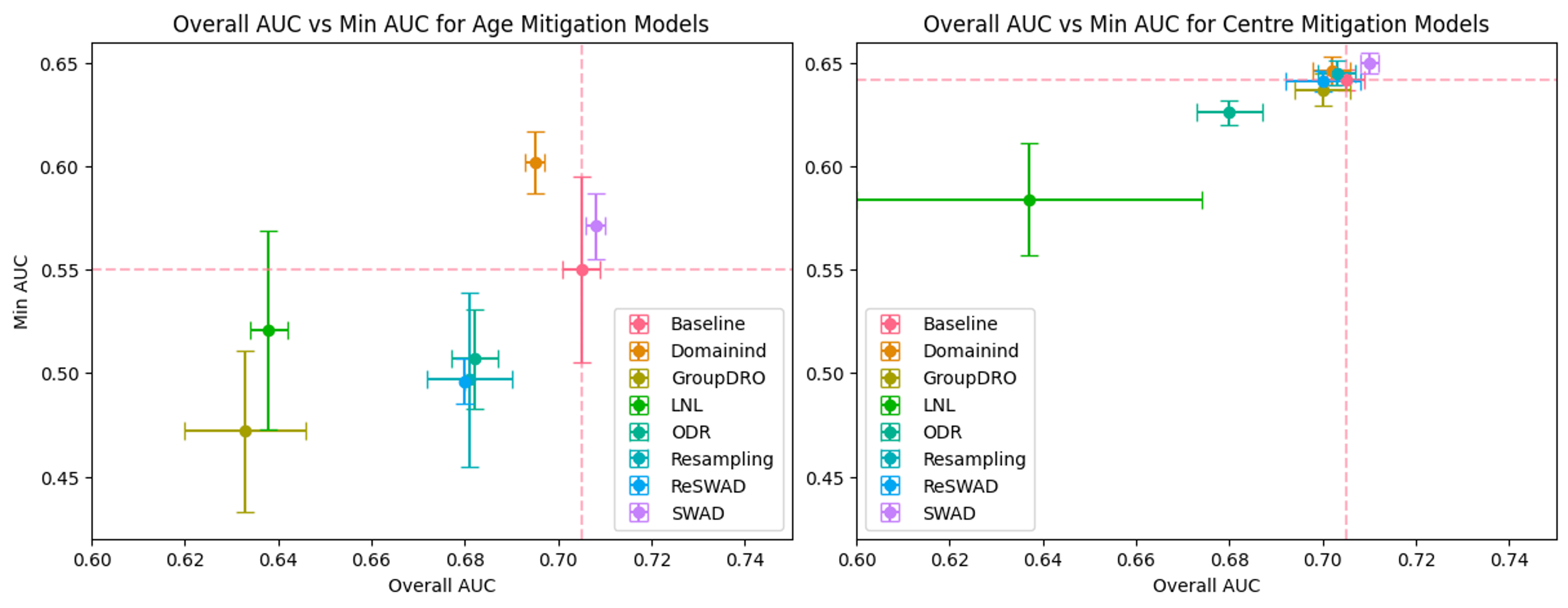}}
    \caption{Overall AUC of age mitigation models (left) and centre mitigation models (right) relative to worst-group AUC. Most models worsen both overall and minimum performance relative to the baseline (red point), especially for age mitigation. Error bars represent standard deviation for 3 random seeds.}
    \label{mit auc min auc plot}
\end{figure}

We then train a number of bias mitigation methods with the objective of reducing the most significant disparities: age, assessment centre, and sex. Initially, we assess how these methods impact overall model performance across all samples, examining whether ``levelling down'' occurs \cite{zietlowLevelingComputerVision2022}. Regarding age mitigation, SWAD is the only method capable of maintaining overall AUC, whereas all other mitigation methods result in a decrease in AUC, particularly gDRO (Figure \ref{mit auc min auc plot}). Interestingly, this decrease in AUC is less pronounced in the assessment centre mitigation models. Only LNL and ODR show a notable decrease in AUC and precision, whereas the other models show similar overall performance across all four metrics (Figure A3). Sex disparity mitigation has a more variable effect (see Figure A4).

\subsubsection{Disparity Reduction}

Overall, no methods achieve their intended effects of reducing disparities and boosting worst group performance. For age-related disparities, DomainInd is the only model which shows some effectiveness; it decreases accuracy, AUC, and recall gap relative to baseline while also increasing worst-group performance (Table \ref{age centre mit disparity table}). However, it also causes a slight reduction in overall performance (Figure A3). SWAD performance is generally similar to baseline performance, but other models decrease min AUC and min precision. 

For centre-related disparities, the effectiveness of the models in improving fairness is very limited, especially in boosting worst-group performance. SWAD is the only method which maintains or slightly improves upon baseline disparities (Table \ref{age centre mit disparity table}). Other methods have negative effects on at least one of the metrics. For instance, resampling increases accuracy gap, ODR lowers min AUC by 0.02, and recalibration lowers min recall by 0.02. We also note that the optimal per-subgroup decision thresholds (for recalibration) range from 0.50 to 0.73, suggesting the baseline model does not uniformly adapt to the characteristics of different subgroups. 

\begin{table}[htbp]
\centering
\caption{Performance disparities across age groups and assessment centres for their respective mitigation models. DomainInd is the only method able to reduce most age disparities relative to the baseline model, while no models are able to consistently reduce assessment centre disparities. Standard deviation of the three random seeds shown in parentheses.}
\resizebox{\columnwidth}{!}{%
\begin{tabular}{
>{\columncolor[HTML]{FFFFFF}}l 
>{\columncolor[HTML]{FFFFFF}}l 
>{\columncolor[HTML]{FFFFFF}}l 
>{\columncolor[HTML]{FFFFFF}}l 
>{\columncolor[HTML]{FFFFFF}}l 
>{\columncolor[HTML]{FFFFFF}}l 
>{\columncolor[HTML]{FFFFFF}}l 
>{\columncolor[HTML]{FFFFFF}}l 
>{\columncolor[HTML]{FFFFFF}}l 
>{\columncolor[HTML]{FFFFFF}}l }
\toprule
                                                          & \textbf{Model}         & \textbf{Acc. Gap}$\downarrow$  & \textbf{Min Acc.}$\uparrow$  & \textbf{AUC Gap}$\downarrow$       & \textbf{Min AUC}$\uparrow$       & \textbf{Prec. Gap}$\downarrow$ & \textbf{Min Prec.}$\uparrow$ & \textbf{Rec. Gap}$\downarrow$    & \textbf{Min Rec.}$\uparrow$    \\ \midrule
\cellcolor[HTML]{FFFFFF}                                  & \textbf{Baseline}      & 0.187 (0.017)          & 0.639 (0.017)          & 0.15 (0.04)            & 0.55 (0.045)           & 0.129 (0.005)          & 0.724 (0.008)          & 0.328 (0.045)          & 0.643 (0.055)          \\ \cline{2-10} 
\cellcolor[HTML]{FFFFFF}                                  & \textbf{DomainInd}     & \textbf{0.158 (0.018)} & 0.658 (0.007)          & \textbf{0.101 (0.013)} & \textbf{0.602 (0.015)} & 0.161 (0.005)          & 0.702 (0.004)          & \textbf{0.189 (0.029)} & 0.743 (0.023)          \\ \cline{2-10} 
\cellcolor[HTML]{FFFFFF}                                  & \textbf{GroupDRO}      & 0.245 (0.014)          & 0.6 (0.014)            & 0.147 (0.018)          & 0.472 (0.039)          & 0.178 (0.012)          & 0.668 (0.012)          & 0.341 (0.019)          & 0.659 (0.019)          \\ \cline{2-10} 
\cellcolor[HTML]{FFFFFF}                                  & \textbf{LNL}           & 0.236 (0.01)           & 0.61 (0.01)            & 0.107 (0.053)          & 0.521 (0.048)          & 0.163 (0.008)          & 0.684 (0.009)          & 0.349 (0.054)          & 0.649 (0.055)          \\ \cline{2-10} 
\cellcolor[HTML]{FFFFFF}                                  & \textbf{ODR}           & 0.184 (0.021)          & 0.647 (0.008)          & 0.174 (0.038)          & 0.507 (0.024)          & 0.142 (0.004)          & 0.705 (0.006)          & 0.277 (0.026)          & 0.704 (0.011)          \\ \cline{2-10} 
\cellcolor[HTML]{FFFFFF}                                  & \textbf{Recalibration} & 0.175 (0.007)          & \textbf{0.664 (0.004)} & 0.15 (0.04)            & 0.55 (0.045)           & 0.156 (0.007)          & 0.699 (0.005)          & 0.22 (0.024)           & \textbf{0.768 (0.013)} \\ \cline{2-10} 
\cellcolor[HTML]{FFFFFF}                                  & \textbf{Resampling}    & 0.181 (0.006)          & 0.647 (0.014)          & 0.185 (0.053)          & 0.497 (0.042)          & 0.136 (0.016)          & 0.71 (0.017)           & 0.283 (0.014)          & 0.692 (0.013)          \\ \cline{2-10} 
\cellcolor[HTML]{FFFFFF}                                  & \textbf{ReSWAD}        & 0.187 (0.027)          & 0.647 (0.018)          & 0.191 (0.013)          & 0.496 (0.011)          & 0.122 (0.007)          & 0.725 (0.006)          & 0.323 (0.056)          & 0.66 (0.048)           \\ \cline{2-10} 
\multirow{-9}{*}{\cellcolor[HTML]{FFFFFF}\textbf{Age}}    & \textbf{SWAD}          & 0.192 (0.003)          & 0.646 (0.003)          & 0.13 (0.018)           & 0.571 (0.016)          & \textbf{0.121 (0.005)} & \textbf{0.729 (0.009)} & 0.339 (0.021)          & 0.65 (0.024)           \\ \hline
\cellcolor[HTML]{FFFFFF}                                  & \textbf{Baseline}      & 0.061 (0.012)          & 0.706 (0.013)          & 0.104 (0.004)          & 0.642 (0.005)          & 0.097 (0.012)          & 0.776 (0.01)           & 0.149 (0.013)          & 0.775 (0.029)          \\ \cline{2-10} 
\cellcolor[HTML]{FFFFFF}                                  & \textbf{DomainInd}     & \textbf{0.055 (0.019)} & 0.712 (0.004)          & 0.106 (0.005)          & 0.646 (0.007)          & 0.111 (0.007)          & 0.763 (0.002)          & 0.189 (0.027)          & 0.78 (0.034)           \\ \cline{2-10} 
\cellcolor[HTML]{FFFFFF}                                  & \textbf{GroupDRO}      & 0.061 (0.006)          & 0.71 (0.001)           & 0.105 (0.003)          & 0.637 (0.008)          & 0.106 (0.002)          & 0.765 (0.003)          & 0.183 (0.009)          & 0.779 (0.014)          \\ \cline{2-10} 
\cellcolor[HTML]{FFFFFF}                                  & \textbf{LNL}           & 0.082 (0.007)          & 0.682 (0.015)          & \textbf{0.089 (0.017)} & 0.584 (0.027)          & \textbf{0.092 (0.004)} & 0.751 (0.01)           & 0.197 (0.071)          & 0.785 (0.088)          \\ \cline{2-10} 
\cellcolor[HTML]{FFFFFF}                                  & \textbf{ODR}           & 0.065 (0.017)          & 0.71 (0.006)           & 0.097 (0.003)          & 0.626 (0.006)          & 0.098 (0.006)          & 0.765 (0.006)          & \textbf{0.137 (0.023)} & \textbf{0.808 (0.015)} \\ \cline{2-10} 
\cellcolor[HTML]{FFFFFF}                                  & \textbf{Recalibration} & 0.118 (0.022)          & 0.711 (0.015)          & 0.104 (0.004)          & 0.642 (0.005)          & 0.079 (0.011)          & \textbf{0.781 (0.007)} & 0.19 (0.071)           & 0.755 (0.05)           \\ \cline{2-10} 
\cellcolor[HTML]{FFFFFF}                                  & \textbf{Resampling}    & 0.085 (0.013)          & 0.712 (0.011)          & 0.098 (0.013)          & 0.645 (0.006)          & 0.101 (0.006)          & 0.769 (0.003)          & 0.146 (0.022)          & 0.799 (0.033)          \\ \cline{2-10} 
\cellcolor[HTML]{FFFFFF}                                  & \textbf{ReSWAD}        & 0.082 (0.008)          & 0.712 (0.008)          & 0.097 (0.005)          & 0.641 (0.005)          & 0.107 (0.005)          & 0.762 (0.006)          & 0.174 (0.019)          & 0.793 (0.012)          \\ \cline{2-10} 
\multirow{-9}{*}{\cellcolor[HTML]{FFFFFF}\textbf{Centre}} & \textbf{SWAD}          & 0.06 (0.019)           & \textbf{0.715 (0.013)} & 0.095 (0.009)          & \textbf{0.65 (0.005)}  & 0.102 (0.007)          & 0.772 (0.009)          & 0.156 (0.016)          & 0.776 (0.046)          \\ \bottomrule
\end{tabular}%
}
\label{age centre mit disparity table}
\end{table}

\section{Conclusions}

Our model trained with retinal images from the UKBB shows notably poor performance on certain subgroups of the population. In particular, although some level of age- or sex-related disparities could be expected due to differences in biological manifestation or prevalence of hypertension, centre disparities (which cannot be explained by any of the investigated confounders), are unexpected given the standardisation of the UKBB. These disparities would lead to unfair outcomes if such a model was deployed. This highlights the importance of systematically conducting a granular assessment of a model's performance. 

Moreover, existing methods largely fail to mitigate these disparities. Most methods, particularly for age disparity mitigation, have a detrimental effect on overall performance. Even worse, few really improve fairness, and while some may show marginal improvement in one scenario, they adversely impact others. For instance, the DomainInd model slightly improves age- and sex-related disparities but does not show improvements in assessment-centre disparities. No method is actually able to boost performance for assessment centre f, suggesting that further methodological advancements are necessary, or that perhaps a maximum performance has already been reached rendering mitigation efforts ineffective. These observations highlight how applying bias mitigation methods indiscriminately may actually worsen overall outcomes and exacerbate existing disparities, concordant with recent findings in MEDFAIR \cite{zongMEDFAIRBENCHMARKINGFAIRNESS2023}. Overall, it appears important to precisely characterise biases and their underlying causes, as this understanding is crucial for informing appropriate mitigation strategies.

Future work should continue to develop a framework to better characterise disparities, for example building off previous work done in \cite{jonesCausalPerspectiveDataset2024,jonesRoleSubgroupSeparability2023}. We consider a very narrow scenario of hypertension prediction from retinal images, but it would be interesting to see how these findings extend to other retinal image tasks and other image modalites. It would also be of interest to conduct a more in-depth exploration of the UKBB dataset specifically, in order to understand the interplay between selection bias, dataset standardisation, and subsequent model biases, and shed light on why some assessment centres showed such disparate performance. Investigations of this kind are increasingly important given the rise in large databases and initiatives like the UKBB, and the need to ensure downstream findings stay as unbiased as possible.

\begin{credits}
\subsubsection{\ackname} 
This research has been conducted using
data from UK Biobank, a major biomedical database, with access provided through application 80521.
This work was supported by the EPSRC grant number EP/S024093/1 and the Centre for Doctoral Training in Sustainable Approaches to Biomedical Science: Responsible and Reproducible Research (SABS: R3) Doctoral Training Centre, University of Oxford and by GE Healthcare.
The computational aspects of this research were supported by the Wellcome Trust Core Award Grant Number 203141/Z/16/Z and the NIHR Oxford BRC. The views expressed are those of the author(s) and not necessarily those of the NHS, the NIHR or the Department of Health.
\subsubsection{\discintname}
None.
\end{credits}

\bibliographystyle{splncs04}
\bibliography{mybibliography}

\clearpage  
\section{Appendix}
\renewcommand{\thefigure}{A\arabic{figure}}
\setcounter{figure}{0} % Reset figure counter
\renewcommand{\thetable}{A\arabic{table}}

\setcounter{table}{0} % Reset figure counter
\vspace{0pt}

\begin{figure}[]
    \centering
    \resizebox{8.5cm}{!}
    {\includegraphics{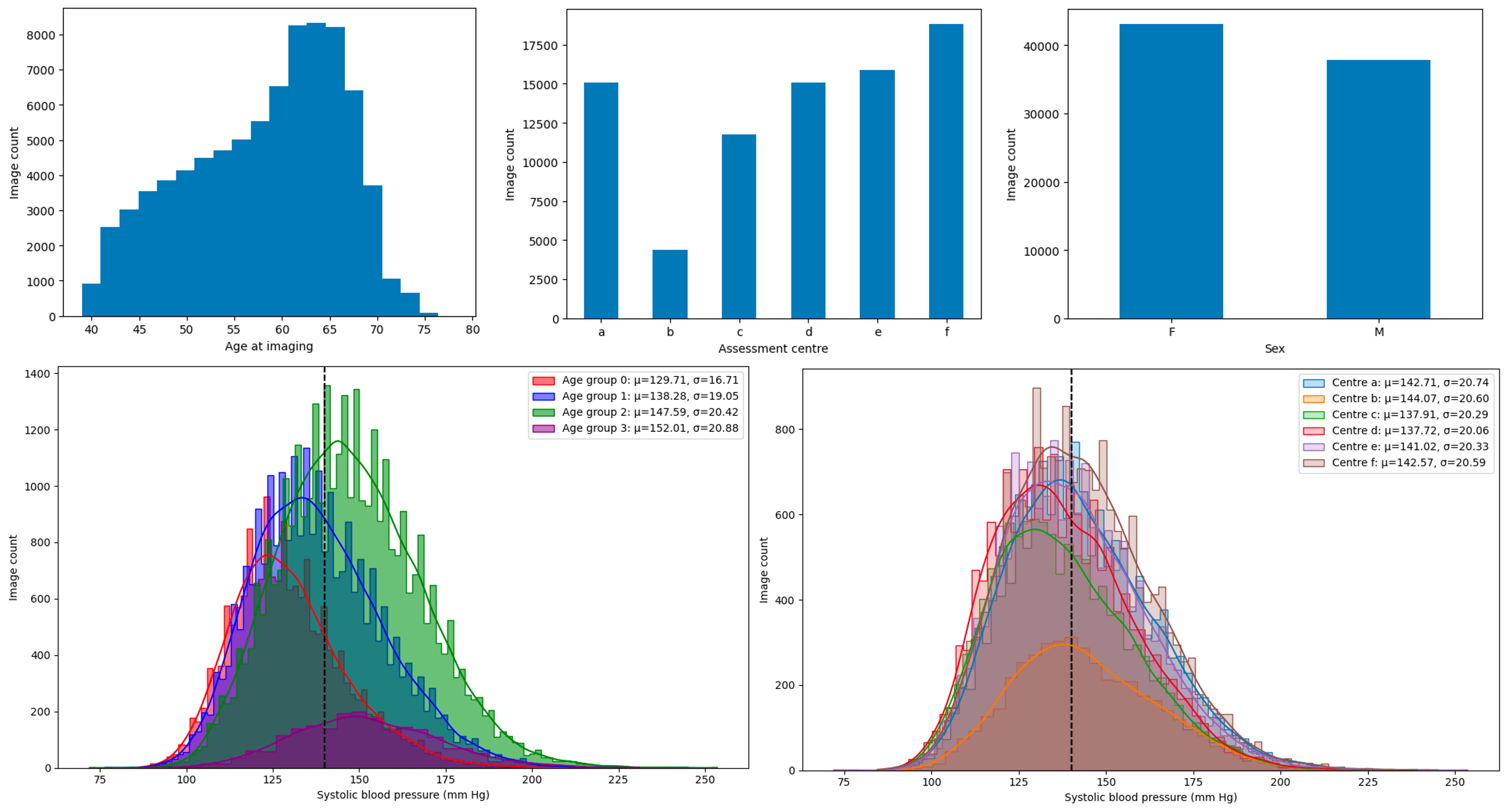}}
    \caption{Baseline data characteristics and SBP distribution.}
    \label{Age and bp dist}
\end{figure}

\vspace{0pt}
\begin{table}[htbp]
\centering
\caption{Baseline model disparities on test set. Standard deviation in parentheses.}
\resizebox{11cm}{!}{%
\begin{tabular}{lllllllllll}
\hline
\textbf{Subgroup}       & \textbf{Accuracy Gap} & \textbf{Min Accuracy} & \textbf{AUC Gap} & \textbf{Min AUC} & \textbf{Precision Gap} & \textbf{Min Precision} & \textbf{Recall Gap} & \textbf{Min Recall} & \textbf{TNR Gap} & \textbf{Min TNR} \\ \hline
\textbf{Age}            & 0.187 (0.017)         & 0.639 (0.017)         & 0.15 (0.04)      & 0.55 (0.045)     & 0.129 (0.005)          & 0.724 (0.008)          & 0.328 (0.045)       & 0.643 (0.055)       & 0.598 (0.033)    & 0.036 (0.031)    \\ \hline
\textbf{Sex}            & 0.062 (0.009)         & 0.701 (0.009)         & 0.033 (0.002)    & 0.676 (0.004)    & 0.067 (0.005)          & 0.78 (0.009)           & 0.068 (0.009)       & 0.794 (0.032)       & 0.135 (0.024)    & 0.356 (0.052)    \\ \hline
\textbf{Alcohol}        & 0.102 (0.005)         & 0.673 (0.017)         & 0.067 (0.009)    & 0.67 (0.01)      & 0.055 (0.009)          & 0.79 (0.012)           & 0.124 (0.015)       & 0.756 (0.043)       & 0.149 (0.02)     & 0.364 (0.038)    \\ \hline
\textbf{Centre}         & 0.061 (0.012)         & 0.706 (0.013)         & 0.104 (0.004)    & 0.642 (0.005)    & 0.097 (0.012)          & 0.776 (0.01)           & 0.149 (0.013)       & 0.775 (0.029)       & 0.357 (0.031)    & 0.219 (0.06)     \\ \hline
\textbf{BMI}            & 0.141 (0.023)         & 0.65 (0.005)          & 0.033 (0.003)    & 0.677 (0.004)    & 0.239 (0.007)          & 0.672 (0.011)          & 0.065 (0.003)       & 0.782 (0.03)        & 0.082 (0.008)    & 0.384 (0.05)     \\ \hline
\textbf{Deprivation}    & 0.019 (0.006)         & 0.722 (0.013)         & 0.015 (0.004)    & 0.698 (0.002)    & 0.024 (0.011)          & 0.802 (0.012)          & 0.019 (0.01)        & 0.821 (0.037)       & 0.045 (0.02)     & 0.422 (0.051)    \\ \hline
\textbf{Ethnicity}      & 0.041 (0.011)         & 0.703 (0.034)         & 0.029 (0.019)    & 0.693 (0.004)    & 0.041 (0.01)           & 0.813 (0.007)          & 0.106 (0.019)       & 0.73 (0.048)        & 0.217 (0.02)     & 0.401 (0.002)    \\ \hline
\textbf{Gen\_ethnicity} & 0.019 (0.011)         & 0.715 (0.018)         & 0.006 (0.004)    & 0.703 (0.003)    & 0.005 (0.004)          & 0.809 (0.008)          & 0.039 (0.012)       & 0.797 (0.036)       & 0.068 (0.008)    & 0.429 (0.047)    \\ \hline
\end{tabular}%
}
\label{Table overall disparities}
\end{table}

\begin{table}[htbp]
\centering
\caption{Assessment centre disparities remain despite conditioning on age group.}
\resizebox{3cm}{!}{%
\begin{tabular}{|
>{\columncolor[HTML]{FFFFFF}}r |
>{\columncolor[HTML]{FFFFFF}}r |
>{\columncolor[HTML]{FFFFFF}}r |
>{\columncolor[HTML]{FFFFFF}}r |
>{\columncolor[HTML]{FFFFFF}}r |
>{\columncolor[HTML]{FFFFFF}}r |
>{\columncolor[HTML]{FFFFFF}}r |}
\hline
{\color[HTML]{000000} \textbf{Age}}                                         & {\color[HTML]{000000} \textbf{Centre}} & {\color[HTML]{000000} \textbf{Accuracy}} & {\color[HTML]{000000} \textbf{AUC}} & {\color[HTML]{000000} \textbf{Precision}} & {\color[HTML]{000000} \textbf{Recall}} & {\color[HTML]{000000} \textbf{TNR}} \\ \hline
\cellcolor[HTML]{FFFFFF}{\color[HTML]{000000} }                             & {\color[HTML]{000000} \textbf{a}}           & {\color[HTML]{000000} 0.647}             & {\color[HTML]{000000} 0.718}        & {\color[HTML]{000000} 0.762}              & {\color[HTML]{000000} 0.622}           & {\color[HTML]{000000} 0.686}        \\ \cline{2-7} 
\cellcolor[HTML]{FFFFFF}{\color[HTML]{000000} }                             & {\color[HTML]{000000} \textbf{b}}           & {\color[HTML]{000000} 0.677}             & {\color[HTML]{000000} 0.740}        & {\color[HTML]{000000} 0.835}              & {\color[HTML]{000000} 0.694}           & {\color[HTML]{000000} 0.630}        \\ \cline{2-7} 
\cellcolor[HTML]{FFFFFF}{\color[HTML]{000000} }                             & {\color[HTML]{000000} \textbf{c}}           & {\color[HTML]{000000} 0.668}             & {\color[HTML]{000000} 0.732}        & {\color[HTML]{000000} 0.745}              & {\color[HTML]{000000} 0.621}           & {\color[HTML]{000000} 0.728}        \\ \cline{2-7} 
\cellcolor[HTML]{FFFFFF}{\color[HTML]{000000} }                             & {\color[HTML]{000000} \textbf{d}}           & {\color[HTML]{000000} 0.622}             & {\color[HTML]{000000} 0.684}        & {\color[HTML]{000000} 0.687}              & {\color[HTML]{000000} 0.593}           & {\color[HTML]{000000} 0.659}        \\ \cline{2-7} 
\cellcolor[HTML]{FFFFFF}{\color[HTML]{000000} }                             & {\color[HTML]{000000} \textbf{e}}           & {\color[HTML]{000000} 0.631}             & {\color[HTML]{000000} 0.710}        & {\color[HTML]{000000} 0.750}              & {\color[HTML]{000000} 0.631}           & {\color[HTML]{000000} 0.630}        \\ \cline{2-7} 
\multirow{-6}{*}{\cellcolor[HTML]{FFFFFF}{\color[HTML]{000000} \textbf{0}}} & {\color[HTML]{000000} \textbf{f}}           & {\color[HTML]{000000} 0.615}             & {\color[HTML]{000000} 0.619}        & {\color[HTML]{000000} 0.624}              & {\color[HTML]{000000} 0.805}           & {\color[HTML]{000000} 0.368}        \\ \hline
\cellcolor[HTML]{FFFFFF}{\color[HTML]{000000} }                             & {\color[HTML]{000000} \textbf{a}}           & {\color[HTML]{000000} 0.725}             & {\color[HTML]{000000} 0.687}        & {\color[HTML]{000000} 0.840}              & {\color[HTML]{000000} 0.798}           & {\color[HTML]{000000} 0.470}        \\ \cline{2-7} 
\cellcolor[HTML]{FFFFFF}{\color[HTML]{000000} }                             & {\color[HTML]{000000} \textbf{b}}           & {\color[HTML]{000000} 0.740}             & {\color[HTML]{000000} 0.712}        & {\color[HTML]{000000} 0.846}              & {\color[HTML]{000000} 0.815}           & {\color[HTML]{000000} 0.476}        \\ \cline{2-7} 
\cellcolor[HTML]{FFFFFF}{\color[HTML]{000000} }                             & {\color[HTML]{000000} \textbf{c}}           & {\color[HTML]{000000} 0.709}             & {\color[HTML]{000000} 0.711}        & {\color[HTML]{000000} 0.791}              & {\color[HTML]{000000} 0.798}           & {\color[HTML]{000000} 0.497}        \\ \cline{2-7} 
\cellcolor[HTML]{FFFFFF}{\color[HTML]{000000} }                             & {\color[HTML]{000000} \textbf{d}}           & {\color[HTML]{000000} 0.695}             & {\color[HTML]{000000} 0.705}        & {\color[HTML]{000000} 0.801}              & {\color[HTML]{000000} 0.761}           & {\color[HTML]{000000} 0.527}        \\ \cline{2-7} 
\cellcolor[HTML]{FFFFFF}{\color[HTML]{000000} }                             & {\color[HTML]{000000} \textbf{e}}           & {\color[HTML]{000000} 0.735}             & {\color[HTML]{000000} 0.717}        & {\color[HTML]{000000} 0.828}              & {\color[HTML]{000000} 0.819}           & {\color[HTML]{000000} 0.476}        \\ \cline{2-7} 
\multirow{-6}{*}{\cellcolor[HTML]{FFFFFF}{\color[HTML]{000000} \textbf{1}}} & {\color[HTML]{000000} \textbf{f}}           & {\color[HTML]{000000} 0.669}             & {\color[HTML]{000000} 0.621}        & {\color[HTML]{000000} 0.696}              & {\color[HTML]{000000} 0.879}           & {\color[HTML]{000000} 0.273}        \\ \hline
\cellcolor[HTML]{FFFFFF}{\color[HTML]{000000} }                             & {\color[HTML]{000000} \textbf{a}}           & {\color[HTML]{000000} 0.786}             & {\color[HTML]{000000} 0.672}        & {\color[HTML]{000000} 0.866}              & {\color[HTML]{000000} 0.884}           & {\color[HTML]{000000} 0.241}        \\ \cline{2-7} 
\cellcolor[HTML]{FFFFFF}{\color[HTML]{000000} }                             & {\color[HTML]{000000} \textbf{b}}           & {\color[HTML]{000000} 0.826}             & {\color[HTML]{000000} 0.638}        & {\color[HTML]{000000} 0.898}              & {\color[HTML]{000000} 0.903}           & {\color[HTML]{000000} 0.282}        \\ \cline{2-7} 
\cellcolor[HTML]{FFFFFF}{\color[HTML]{000000} }                             & {\color[HTML]{000000} \textbf{c}}           & {\color[HTML]{000000} 0.767}             & {\color[HTML]{000000} 0.729}        & {\color[HTML]{000000} 0.872}              & {\color[HTML]{000000} 0.842}           & {\color[HTML]{000000} 0.412}        \\ \cline{2-7} 
\cellcolor[HTML]{FFFFFF}{\color[HTML]{000000} }                             & {\color[HTML]{000000} \textbf{d}}           & {\color[HTML]{000000} 0.767}             & {\color[HTML]{000000} 0.650}        & {\color[HTML]{000000} 0.848}              & {\color[HTML]{000000} 0.874}           & {\color[HTML]{000000} 0.278}        \\ \cline{2-7} 
\cellcolor[HTML]{FFFFFF}{\color[HTML]{000000} }                             & {\color[HTML]{000000} \textbf{e}}           & {\color[HTML]{000000} 0.796}             & {\color[HTML]{000000} 0.691}        & {\color[HTML]{000000} 0.868}              & {\color[HTML]{000000} 0.889}           & {\color[HTML]{000000} 0.330}        \\ \cline{2-7} 
\multirow{-6}{*}{\cellcolor[HTML]{FFFFFF}{\color[HTML]{000000} \textbf{2}}} & {\color[HTML]{000000} \textbf{f}}           & {\color[HTML]{000000} 0.789}             & {\color[HTML]{000000} 0.638}        & {\color[HTML]{000000} 0.818}              & {\color[HTML]{000000} 0.949}           & {\color[HTML]{000000} 0.139}        \\ \hline
{\color[HTML]{000000} \textbf{3}}                                           & {\color[HTML]{000000} \textbf{f}}           & {\color[HTML]{000000} 0.827}             & {\color[HTML]{000000} 0.550}        & {\color[HTML]{000000} 0.847}              & {\color[HTML]{000000} 0.970}           & {\color[HTML]{000000} 0.036}        \\ \hline
\end{tabular}%
}
\label{Table centre condition age}
\end{table}

\begin{figure}[htbp]
    \centering
    \resizebox{12cm}{!}
    {\includegraphics{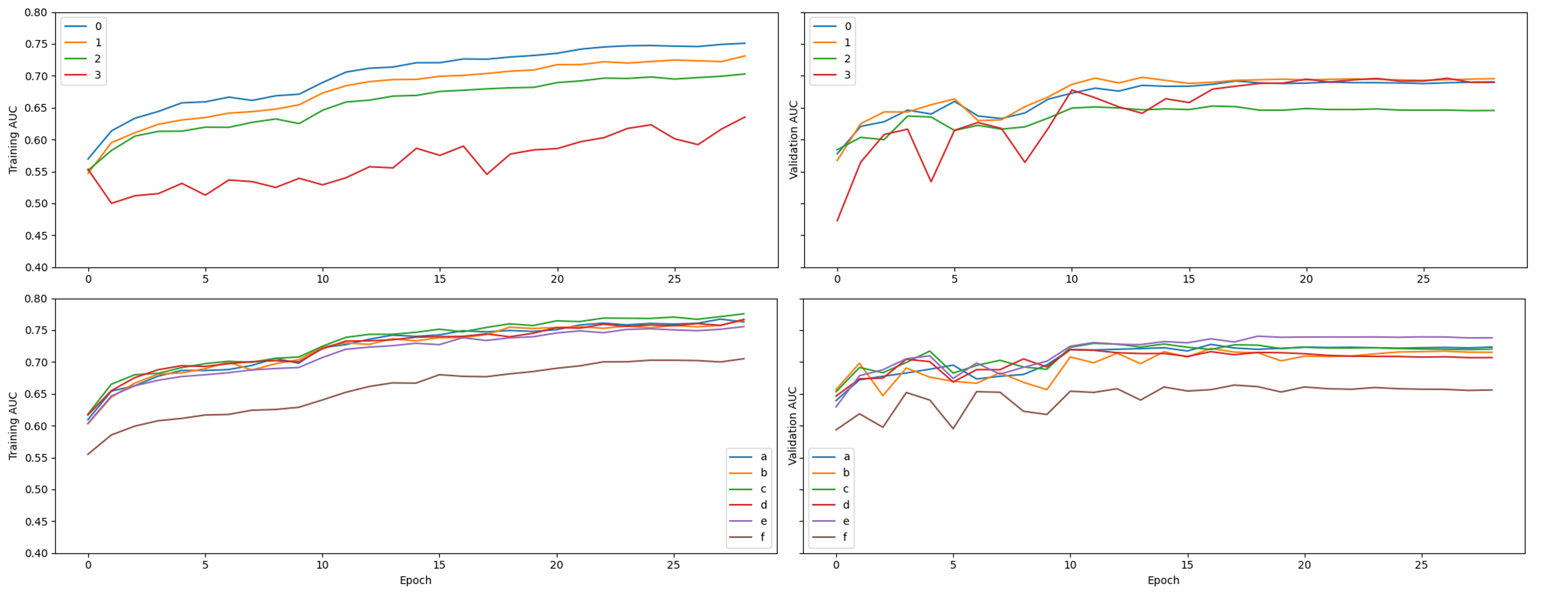}}
    \caption{Age (top) and centre (bottom) AUC evolution during a baseline training run.}
    \label{Evolution age disparities}
\end{figure}

\begin{figure}[htbp]
    \centering
    \resizebox{10cm}{!}{\includegraphics{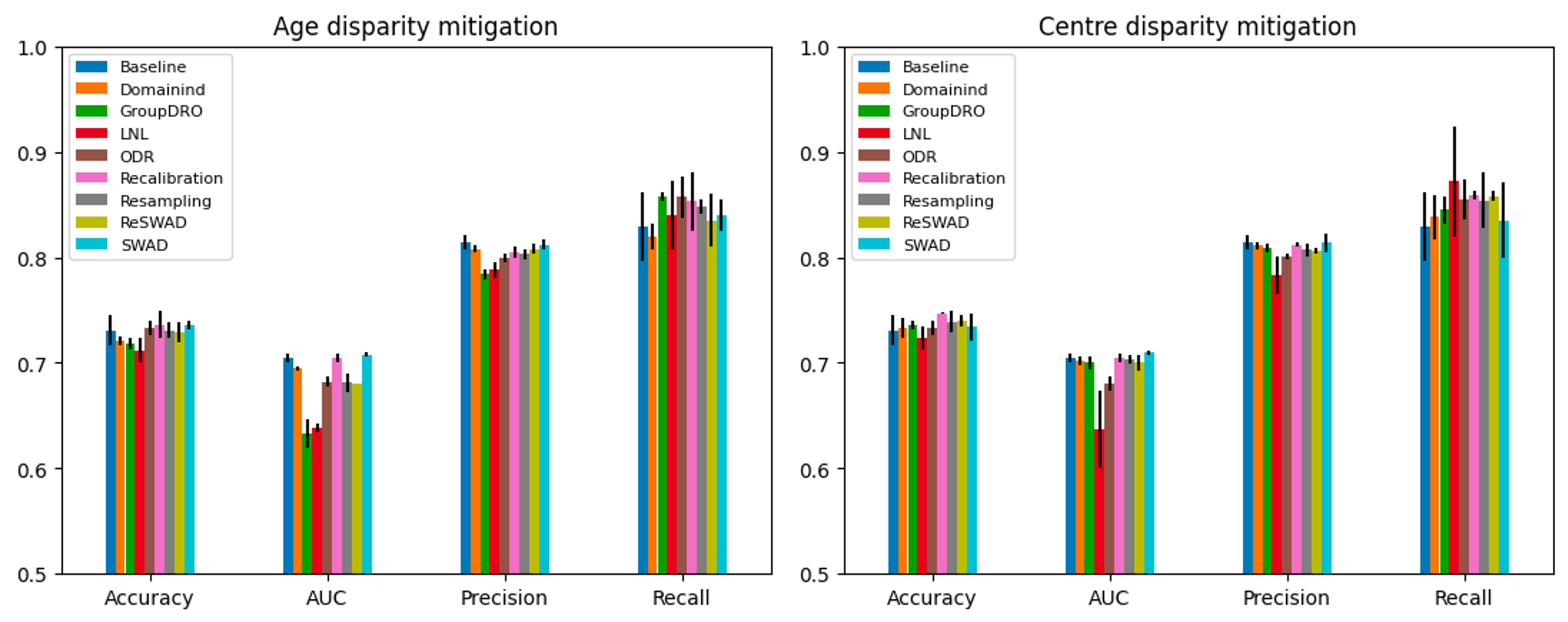}}
    \caption{Overall performance of age mitigation models (left) and centre mitigation models (right). Error bars represent standard deviation for 3 random seeds.}
    \label{mit models disparities plot table}
\end{figure}
% sex disparities
\begin{figure}[htbp]
    \centering
    \resizebox{8cm}{!}{\includegraphics{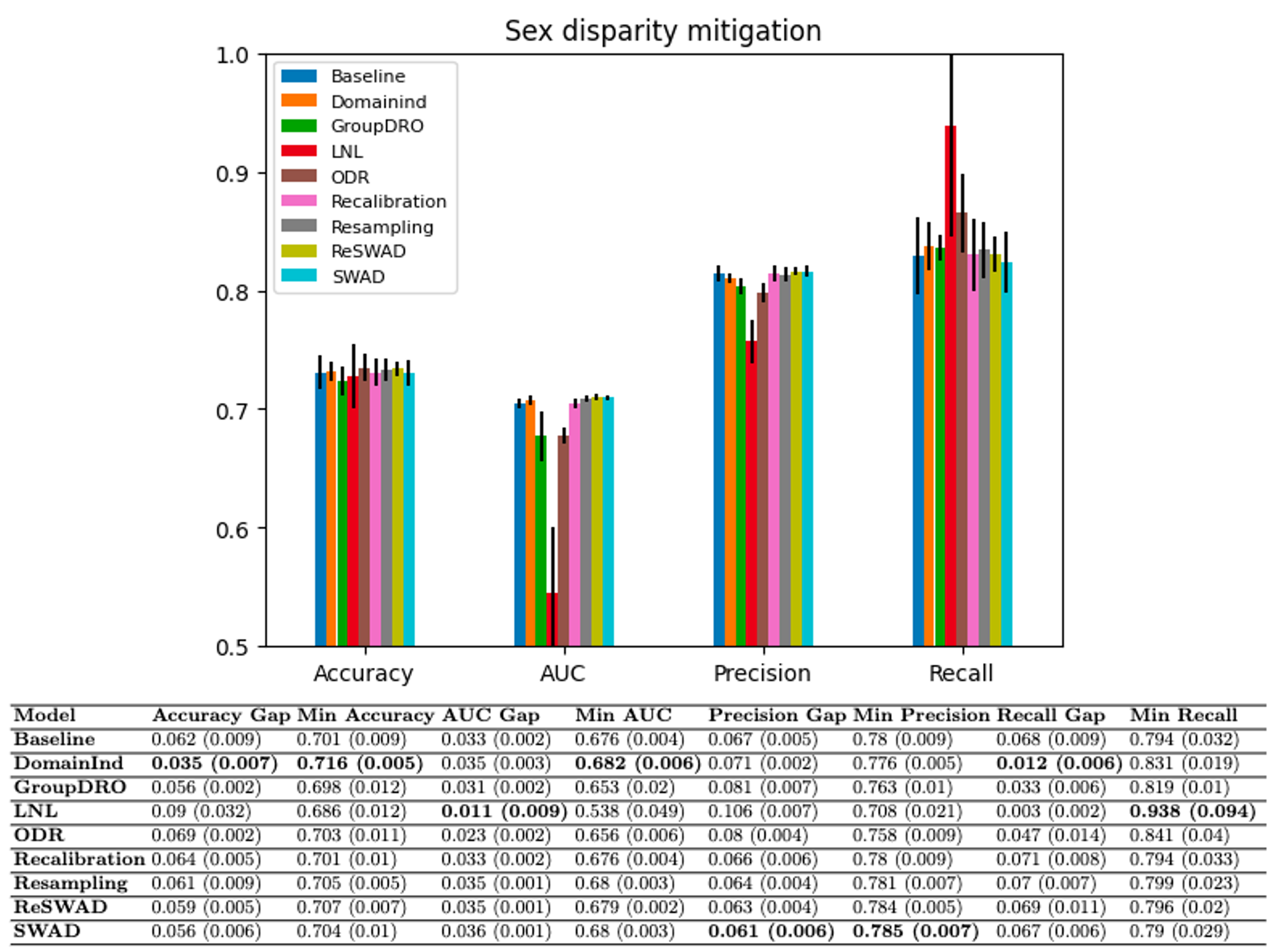}}
    \caption{Overall performance (top) and disparities (bottom) of sex mitigation models.}
    \label{sex disparities overall plot}
\end{figure}

\end{document}